# Application of Machine Learning for Online Reputation Systems

Ahmad Alqwadri    Mohammad Azzeh    Fadi Almasalha

Faculty of Information Technology, Applied Science Private University, Amman 11931, Jordan

**Abstract:** Users on the internet usually require venues to provide better purchasing recommendations. This can be provided by a reputation system that processes ratings to provide recommendations. The rating aggregation process is a main part of reputation system to produce global opinion about the product quality. Naïve methods that are frequently used do not consider consumer profiles in its calculation and cannot discover unfair ratings and trends emerging in new ratings. Other sophisticated rating aggregation methods that use weighted average technique focus on one or a few aspects of consumers' profile data. This paper proposes a new reputation system using machine learning to predict reliability of consumers from consumer profile. In particular, we construct a new consumer profile dataset by extracting a set of factors that have great impact on consumer reliability, which serve as an input to machine learning algorithms. The predicted weight is then integrated with a weighted average method to compute product reputation score. The proposed model has been evaluated over three MovieLens benchmarking datasets, using 10-Folds cross validation. Furthermore, the performance of the proposed model has been compared to previous published rating aggregation models. The obtained results were promising which suggest that the proposed approach could be a potential solution for reputation systems. The results of comparison demonstrated the accuracy of our models. Finally, the proposed approach can be integrated with online recommendation systems to provide better purchasing recommendations and facilitate user experience on online shopping markets.

## 1 Introduction

Online rating is a common venue for consumers to meet their demand when choosing products in online shopping markets [1][2]. Consumers feel confident in expressing their opinions through ratings [3]. Reputation system is an intrinsic part of recommender systems, which can facilitate product choice decision by reflecting global opinion about product [4][5]. The process of aggregating reputation scores for online products is important part of reputation system because it affects choices of consumers, thus targeting consumers' satisfaction [6][7]. The use of reputation systems is increasingly noticed because they are free, widely available, easy to reach, and can facilitate consumer decision [8][9]. The accuracy of computing product reputation score has great influence on the consumer decision because it reflects global opinion about product. In literature, there are too many published reputation systems [8]. Amongst them, the Naive methods (i.e. average and median of ratings) are the frequent methods to compute product quality because they are simple and easy to apply without additional configuration setup. But these methods do not take in consideration the consumers profiles' data in their process or even the popularity of product [10]. It also cannot discover unfair ratings and trend emerging from recent consumer ratings [1] [11]. Therefore, other probabilistic and statistical methods were emerged to handle these limitations [1], [4], [9], [12], [13]. These methods showed good accuracy, but they have large space of configuration possibilities. The weighted average methods are the common alternative to compute the product reputation score, where the weights are measured from different sources such as reliability of consumers [14][15], trust [16][17], leniency of consumer [13] or rating age [18]. The weighted average methods initially require computing quality of consumers' ratings before calculating product score, and then follow predefined threshold built by the expert. These weighted methods require sophisticated processing to obtain reputation score of products. For example, Lenient Quality (LQ) [13] model calculates the weight based on reviews leniency or strictness in providing ratings. However, majority of current weighted methods focus on a single aspect of consumer' ratings such as time of ratings, malicious ratings, or tendency of consumers' ratings. Also, they measure weights form consumer' profile, but they do not predict them.

In summary, we can notice that most of the previous rating aggregation models focus on a few aspects of consumer data. In addition, the machine learning algorithms have not been used intensively during the rating aggregation process to predict weights from consumers' profile data instead of statistical methods. Therefore, this paper proposes a new weighted average approach to compute product reputation score, where weights are predicted from consumers' profile, using machine learning algorithms. To facilitate that, various consumers related variables are extracted from the raw rating dataset, including:

1. *Consumer tendency*, which measures the user behavior in providing ratings, which is expressed by three variables (Number of positive ratings, number of neutral ratings and number of negative ratings given by a consumer).

2. *Consumer fluctuation*, which measures of the variance of consumer ratings from the ratings provided by community.

3. *Consumer experience*, which is the ratio of number of ratings provided by each consumer to the total number of ratings in the system.

4. *Consumer reliability*, which measures the average of errors for all ratings provided by a consumer. This variable shows the reliability of consumer in providing ratings, which measures the closeness of consumer ratings to the average products rating.

The extracted dataset represents description of consumers' ratings where each row represents a consumer data whereas the columns represent the extracted variables. The extracted dataset is entered to machine learning algorithm to predict consumer reliability as a form of weight. The tendency variables in addition to fluctuation and experience variables are considered input variables while reliability is considered as output variable. Multiple machine learning algorithms are used in this paper including, Linear Regression (LR), Support Vector Regression (SVR), K-Nearest Neighbor (KNN) and Regression Tree (RT). The predicted consumer reliability is treated as consumer weight and used with weighted average method to compute the final product quality score. The main research questions that we address in this paper are:

**RQ1**: Does the extract variables have great effect on computing consumer reliability?

**RQ2**: Does using machine learning enables us to compute consumer reliability efficiently and thus enhance accuracy of rating aggregation?

**RQ3**: Which machine learning method can produce better performance?

To answer RQ1, we propose various variables that reflect consumer tendency, experience, and fluctuation in providing ratings. To answer RQ2 and RQ3, we develop four machine learning algorithms to predict consumer reliability. The accuracy of each algorithm is compared to previous reputation systems in order to determine the performance and stability of our proposed model.

The paper is structured as follows: section 2 presents the related work. Section 3 presents the choice of learning methods. Section 4 presents the used Dataset. Section 5 introduces the proposed reputation systems. Section 6 presents evaluation measures. Section 7 described evaluation measures. Section 8 presents results, and finally section 9 ends with conclusion.

## 2 Related Work

Naïve methods are the most frequent used methods for computing ratings in most E-commerce systems [19][20]. The later methods are not informative as they cannot discover recent rating trend and easily influenced by unfair ratings [1], [11]. On the other hand, the weighted average methods work more efficiently than Naïve methods as they consider the consumer data in computing reputation product score. Josang et al. [9] stated that the ratings age is a good factor which can reflect the importance of old or recent ratings. They demonstrated that linear and nonlinear aging discount functions can be used through weighted average method. This technique needs involving professional expert to specifying the unit of age (i.e. day, week, month and year). A different study suggests using the number of past transactions instead of ratings age [4]. Leberknight et al. [10] demonstrated that a higher weight must be given for recent ratings, and the reputation system should take that as well the discounting factor during ratings computation. They proposed a model that divides rating into number of non-overlapping equal subsets, and then investigate the volatility in each subset with respect to the near subset. Finally, the variabilities in all subsets are fused together through discounting function that is used later to compute product score.

Other studies measured weights from consumer data such as reliability, credibility and trust of consumers. Lauw et al. [13] proposed to use leniency ad strictness of consumers in providing ratings. Lenient consumers are those who frequently provide positive ratings regardless of the actual product quality. Strict consumers are those who frequently provide negative ratings regardless of the actual product quality. Jøsang et al. [7] proposed a reputation system based on multinomial Dirichlet probability distribution. Bharadwaj et al. [3] developed some new variables based on work of Jøsang et al. [21] and using fuzzy logic to compute trust of consumer and reputation of product. Cho et al. [22] used three variables to evaluate the reliability of consumer, namely: consumer expertise in a specific category, consumer trust, and co-orientation. These factors are fused together using either arithmetic average, harmonic average, or multiplication. In the same direction, Liu et al. [11] proposed a set of variables to address the problem of unfair ratings, which are fused together using fuzzy logic. The model has been validated using single and multiple attacks procedures. In the same direction, Rezvani et al. [1] proposed a new method to detect unfair rating using randomized algorithm. On the other hand, Abdel-Hafez et al. [4][12] used Beta distribution function for sparse and sense datasets to efficiently compute reputation score for none-popular items. Azzeh et al. [6], [8] proposed two reputation systems where the first one based on moving average and the second one is based on Fuzzy logic. The first approach assumes to measure variability of data within a window, that is determined based on specified thresholds, then reflect the variability to weight. Regarding Fuzzy Logic, Azzeh et al. [8] proposed four factors from consumer profile that serve as input for predefined Fuzzy Logic System to measure consumer influence.

Other studies focused on examining various factors that affect reputation systems [23-31]. particularly, Wu et al. [23] examined the impact of initial configuration on identifying online user reputation for the user–object bipartite networks. They employed multiple datasets from two sources: Netflix and MoviLens. The results showed that the Online users' reputations increase as users rate more and more items. Yang et al. [24] found that online ratings are subject to anchoring bias where users tend to give a low rating after low rating and high rating after high rating. Gao et al. [25] proposed group-based ranking method to evaluate users' reputations based on their grouping behaviors. This can support reputation system and online rating ranking. They found that their proposed model is more accurate than correlation method in the presence of spamming attacks. Chen et al. [26] proposed a trust-based recommendation method after integrating the information of trust relations into the resource-redistribution process. they involved a tunable parameter to scale the resources received by trusted users before the redistribution back to the objects. From these studies we can notice that none of them applied machine learning to predict user reliability from user profile data, which is the main objective of this paper.

## 3 Choice of Machine Learning Algorithms

In this study, four common machine learning regression algorithms are used by reason of the good and stable performance in different fields. These algorithms are Linear Regression (LR), Support Vector Regression (SVR), K-Nearest Neighbor (KNN) and Regression Tree (RT). SVR is supervised machine learning algorithm that is used to predict both linear and non-linear output. The SVR is controlled by many tuning parameters which have significant impact on its accuracy. These parameters are: 1) type of kernel function, 2) hyperplane construction method. The SVR attempts to find the optimal hyperplane (Margin), which is the maximum distance between the linear model and the "support points" close to the decision boundary. If there are no points near Margin, then the derived hyperplane can perfectly separate the data with minimum error.

RT is another supervised machine learning algorithm used to predict the continuous value. The algorithm uses Gini or Entropy variable for identifying the optimal divisible features. This process is named as binary recursive partitioning, which continuously split data into small subsets of data and stop when the algorithm cannot divide data into more coherent groups. Finally, the average of output in each leaf node is considered as representative point for the group.

LR is supervised machine learning algorithm used to predict the continuous values. There are two types of this algorithm, the simple linear regression that uses one value of input to predict output with continuous values in constant slope, and the multiple linear regression that uses more than one value of input to predict output. To perfectly constructing a linear model, all input variables must be checked against normal distribution, in case if the input variable does not meet this condition then it is transformed to another scale using logarithmic function.

KNN is a machine learning algorithm that uses similarity measures to retrieve the closest data points to the new case. The algorithm requires determining the number of nearest neighbors (k) and weighting mechanism if necessary, before running algorithm. The Euclidean distance is usually used as similarity measure to identify nearest observations.

## 4 The Proposed Rating Aggregation Method

To evaluate the proposed model, we used three variants of MovieLens datasets [32]. Each dataset has different number of consumer ratings for Movies. We use three types of datasets to evaluate our proposed model as shown in Table 1. The first dataset is called 100K which consist of 943 consumers that rated 1682 movies, and the total rating count is 100,000. The second dataset is called 1M which consists of 6040 consumers and 3706 movies including 1,000,209 ratings count. The third dataset is called 10M that consists of 71,567 consumers, 10,681 movies, and the total count of ratings is 10,000,054. As shown in the Table 2, each MovieLens dataset contains the following attributes 1) ConsumerID 2) MovieID 3) Rating in range 0 to 5, and finally 4) Timestamp which is measured using Unix time.

Table 1. Description of MovieLens Datasets

| Dataset | Consumer Count | Movie Count | Total Rating Count |
|---|---|---|---|
| 100K | 934 | 1682 | 100,000 |
| 1M | 6040 | 3706 | 1,000,209 |
| 10M | 71,567 | 10,681 | 10,000,054 |

Table 2. MovieLens Dataset Summary

| Attributes | Type | Description |
|---|---|---|
| ConsumerID | Numeric (1 -6040) | Consumer ID |
| MovieID | Numeric(1-3952) | Movie ID |
| Rating | Numeric (1-5) | Rating of the Movie |
| Timestamp | Numeric (Unix time) | Time of rating in second |

## 5 The Proposed Rating Aggregation Method

In this paper, we propose a new weighted average reputation aggregation model which uses machine learning as core module to predict consumers' weight as part of computing product reputation score. The general reputation system that is used to compute the product reputation score is described in equation 1.

$$score_i = \frac{\sum w_j * r_j}{\sum w_j} \quad (1)$$

Where $w_j$ is the predicted weight for consumer, *j* who rated product and *i* with rating value $r_j$.

The machine learning algorithms here are used to predict the weight of each consumer. To facilitate that, the raw dataset is processed from the current form to a proper input form; therefore, a set of new variables are extracted from the raw rating dataset which describe the characteristics of each consumer. We believe that these variables can help in predicting the weight of each consumer. The extracted variables are:

**- Consumer Tendency** measures the strictness and leniency of consumer in providing rating. This factor can be measured by three variables (number of positive ratings (pos), number of neutral ratings (nut) and number of negative ratings (ngv)). The Positive variable counts the number of positive ratings that fall in range [4 to 5]. The neutral variable counts the number of neutral rating that equals to 3. Finally, the negative variable counts the numbers of negative ratings that fall in the range [1 to 2].

**- Fluctuation** measures how much far the rating given by a consumer from other consumers for that product. This variable can be formulated as discounting function as shown in equation 2. If the consumer under investigation provided ratings close to other consumers over all shared products, then s/he gets a fluctuation value close to one. Otherwise the value will be discounted according to amount of differences.

$$fluc_i = \frac{1}{m}\sum_{k=1}^{m} \frac{1}{n}\sum_{j}^{n} \lambda^{|r_{ik}-r_{jk}|} \quad (2)$$

Where $n$ is the number of consumers. The $\lambda$ is the fading variable that is used as discounting factor which in our case we use $\lambda=0.95$. m is the number of shared products between the consumer $i$ and other consumers. $r_{ik}$ is the rating given by consumer $i$ for product $k$, while $r_{jk}$ is the rating given by consumer $j$ for product k.

**- Experience** measures the ratio of rating given by a consumer $i$ from the total rating given by all consumer in the raw dataset to see the experience of consumer in providing ratings. The higher the number the better the experience. The reviewer's experience is very important in determining the reviewer's confidence and his ability to provide true ratings. This factor can be assessed by finding the ratio between number of ratings provided by reviewer $u_i$ and maximum reviewer ratings in the dataset, as shown in equation 3.

$$f_{i4} = \frac{|u_i|}{max\{|u_1|,|u_2|,|u_3|,...,|u_n|\}} \quad (3)$$

Where $|u_i|$ is the number of ratings given by a consumer $i$.

**- Reliability** measures the average of errors for all ratings given by a consumer $i$. For each consumer, we calculate the difference between its ratings and the products average ratings. The obtained errors are then averaged to compute the trustworthiness. This factor will be used as consumer weight (i.e. the output variable when using machine learning methods), and we can calculate the consumer weights as shown in equation 4.

$$rel_i = \frac{1}{m}\sum_{k=1}^{m} |r_{ik} - \bar{r_k}| \quad (4)$$

Where $r_k$ is the average of ratings for product $k$.

The summary of all extracted variables is shown in Table 3. The above six variables are collected for each consumer from raw rating dataset to form a new consumer profile dataset. The consumer profile dataset is used to learn weight of consumer through machine learning algorithm as shown in Fig. 1. All variables in the dataset will be used as input, except reliability variable will be served as output. The four employed machine learning algorithms (SVM, RT, LR and KNN) will be used to build prediction models. These models will be validated using 10-Fold Cross validation. In each iteration 90% of the data will be used as training while the remaining data is served as testing. This process is repeated 10 times until all data are tested. The error values are recorded in each iteration then they are averaged to obtain final error. After that, we calculate the product scores for each product as shown in equation 1.

Table 3. Description of consumer profile dataset.

| Variable | Type |
|---|---|
| Positive Rating Count (pos) | Numeric |
| Neutral Rating Count (nut) | Numeric |
| Negative Rating Count (ng) | Numeric |
| Experience(exp) | Numeric |
| Fluctuation (fluc) | Numeric |
| Reliability (rel) | Numeric |

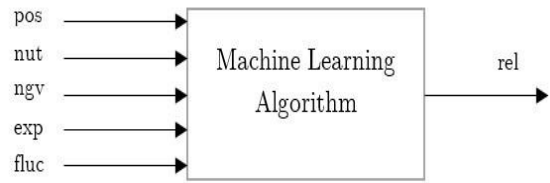

Fig. 1 Machine learning module that is used to predict consumers' reliability.

## 6 Evaluation Measures

In literature, there is no agreed evaluation measure to validate reputation systems; however, we will use the common measures that are used by previous researches [4], [8]. First, we use Mean Absolute Error (MAE) that calculates how much the predicted score are closed to actual ratings for the product. To find the MAE of all cases we calculate the difference between the actual rating and predicted rating as shown in equation 5.

$$MAE = \frac{1}{m}\sum_{k=1}^{m}\frac{\sum_{i=1}^{n}(r_{ik} - score_k)}{n} \quad (5)$$

Where $score_k$ is the generated score for product $k$. $m$ is the number of products in the testing data. $n$ is the number of ratings for $k^{th}$ product in the testing data.

There is another evaluation measure called Kendall Tau coefficient which finds the correlation between two ranked list. The outcome of this analysis is a value between -1 and +1. If the later calculated value is close to -1 then it represents a total disagreement while it represents a total agreement if the value is closer to +1. If the value is close to zero, then that means there is no agreement at all. In our case, the good results are achieved when two list have different rankings which confirms that both reputation systems are different. To investigate the sensitivity of this analysis, we compute the similarity over a specified percentage of the top ranked product. We have chosen 10%, 20%....100% as threshold points. The main objective of this analysis is that the consumers are usually concerned about top products, and to confirm that our model produces relatively different list of ranked products from other models.

## 7 Research Methodology

As we mentioned before in section 4 there are six new variables that have been extracted from the rating raw dataset. All variables are supposed to be normalized in order to have the same influence. We will use Min-Max scaling technique to transfer all variables into scale 0 to 1. These variables form the input and output to the employed machine learning methods in order to predict consumer weight from reliability variable. In the first step of our empirical evaluation, we divided consumers' profile dataset into groups of training and testing sets using 10-folds cross validation. In each validation step, the training dataset (90% of the entire data) is used to learn the machine learning model while the testing data (10% of the entire data) is used for consumer reliability prediction. This procedure is repeated ten times until all testing subsets are validated. The predicted weight for each consumer is stored to be used later when computing product reputation score as explained in equation 1. The accuracy of this procedure is assessed using MAE and Kendall tau correlation as discussed in section 5.

The entire experiments were designed and implemented using Python. From Python we used the following libraries: Pandas to import the dataset, DataFrame to access the dataset as a data frame on python, Itertools to access all data on data frame loop, csv to access the dataset and to create a new csv file from extracted factors, Numpy to deal with numbers, Sklearn to use the machine learning algorithms and mean absolute error, mysql connector to connect and access the database on MySQL and finally Scipy to use Kendall Tau coefficient. Also, we use MySQL 7.3.12 to store the extracted variables from original dataset, execute the SQL operations that handles the consumer weights, and to find the product scores for each product (actual product scores).

The parameter configuration for each kind of machine learning algorithm is described here. For KNN, we set nearest neighbor k=5 to avoid bias, and Euclidean distance as similarity measure. For SVR we used Radial basis Function as kernel function and gamma with auto value. For LR, we checked that variables whether respect normal distribution, if not we transform it into another scale using algorithmic function, we also set random state = 0. Finally, for RT we used categorical/regression algorithm (CART) for building the prediction model. The constructed models are also compared to previous reputation systems that already published in literature. Strictly speaking, we compare our model to the following previous reputation systems such as Average, Median, BetaDR [4], Bayesian [33], Dirichlet [21], IMDb, Fuzzy [11], and LQ [13].

## 8 Results

This section presents the results of our constructed models, in addition to the comparison with other known reputation systems mentioned before in section 6. The MAE evaluation measure was used to assess the accuracy of reputation systems by assessing the differences between actual products scores and their predicted scores. Note that the machine learning models are used only to predict consumer weight from reliability variable, then the weighted average method is used to compute the final product reputation score. The results of MAE for all reputation systems are computed after calculating product reputation scores, which are presented in Table 4. We can notice that all results, over all data sets, are quite small which means that our reputation systems have capability to predict the correct weight for each consumer based on its provided ratings. Amongst them, RT surpasses other models because it has capability to classify data into more coherent groups for which the consumer weight is predicted from closest consumers. Surpassingly, the LR model beats KNN even though, most recommender systems favor KNN because it can identify closest consumers based on the idea of matching. However, the differences among the four machine leaning algorithms are not significant. The second important observation is the stability of results over all datasets. We can notice that RT is the superior over all datasets, followed by LR then by KNN and SVR respectively. This stability is important factor in identifying the most accurate models.

Table 4. MAE accuracy values of the four reputation models

| Dataset | LR | RT | SVR | KNN |
|---------|------|------|------|------|
| 100K    | 0.75 | 0.71 | 0.82 | 0.79 |
| 1M      | 0.73 | 0.69 | 0.77 | 0.76 |
| 10M     | 0.67 | 0.65 | 0.78 | 0.72 |

Table 5 shows the results of previous reputation systems from literature. We followed the same validation procedure conducted over our models with previous reputation systems. Particularly, the Average and Median models do not require to undergo the cross-validation procedure because they do not involve consumer weight computation. The remaining models are undergone to the same 10-Folds cross validation. Note here, these models measure but not predict the consumer weights from raw data. This is the main difference between our approach and previous approaches. We can observe that none of the previous models has beaten our results, therefore we can confirm that our proposed procedure is more accurate than previous models' procedures. Hence, we can notice that our models give best accuracy in comparison with other models over sparse dataset and dense dataset. Surprisingly, the naïve median method outperforms all sophisticated weighted average methods. This might confirm that the naïve method is still useful in some domains, but further investigation is still needed to see if this is true for other domains. The good news from this comparison is that our ML models give higher accuracy than naive models (Average, Median), also the comparison with Bayesian and Fuzzy models gives more accuracy. In addition, our model gives higher accuracy than commercial reputation system like IMDb and in 1M dataset we noticed that our ML models give higher accuracy compared with other reputation systems.

Table 5. Comparison with previous using MAE evaluation measures.

| Dataset | Average | Median | BetaDR | Bayesian | Dirichlet | IMDb | Fuzzy | LQ |
|---|---|---|---|---|---|---|---|---|
| 100K | 0.91 | 0.89 | 0.89 | 0.90 | 0.89 | 0.91 | 0.92 | 1.02 |
| 1M | 0.86 | 0.84 | 0.84 | 0.86 | 0.84 | 0.87 | 0.87 | 0.97 |
| 10M | 0.84 | 0.81 | 0.83 | 0.84 | 0.84 | 0.86 | 0.85 | 0.96 |

To investigate the stability of all reputation systems, we rank the four machine learning models and previous reputation systems based on their MAE values as shown in Table 6. We can notice that RT model is ranked first with high accuracy and the LQ is the lower accuracy. Notably, we can see a stable ranking for all models across all datasets despite slight rank changes for some models like Average and Bayesian.

In addition to the above analysis we performed Kendall tau correlation to compare between two different ranked lists. The main objective of this analysis is to confirm that our model produces relatively different list of top ranked products from other models because the consumers are usually concerned about top products. The good results are obtained when two lists have different rankings which confirm that both reputation systems are different. To investigate the sensitivity of this analysis, we compute the similarity over a specified percentage of the top ranked product. We have chosen 1%, 10%, and 20%,30%....100% as threshold points. In other words, we rank the top products based on their predicted scores, then we chose each time a threshold like 10%. For those selected products we compute Kendall tau coefficient. This process is repeated but for other set of thresholds (i.e. 20%, 30% to 100%). Figures 2 to 5 summarize the Kendal tau sensitivity analysis, where each figure shows a comparison between one of our reputation systems and previous published models over a specified dataset. The horizontal axis represents percentage of top products and the vertical axis represents the Kendall tau values. The main observation that is found from these figures is that there is common trend in all comparisons. They begin with perfect agreement or disagreement and start declining to reach a level near to zero which indicates no similarity between two ranked lists. These results confirm that our reputation systems produce relatively different top ranked list than other model, which necessarily demonstrate that our models are significantly different in computing products reputation scores.

Fig. 2 shows comparison between LR reputation model and other models over three datasets. For 100K dataset as shown in Fig. 2(a), it is noticed that our model ranks 1% of top product quite similarly to Median, Fuzzy, and BetaDR models. However, the correlation degree began to decline after using top 10%. The same trend is observed for 1M as shown in Fig. 2(b) where our model shows relatively small similarity degree with other models, specifically Fuzzy model, at 10% which ranks top products differently from our models at various percentages of top products. Notably, our model and LQ, BetaDR and average models rank top products differently, which indicates that our model is more accurate as confirmed by MAE. For Large dataset 10M, we can notice that our model produces quite similar top product list to Fuzzy, BetaDR, and Bayesian when we look at top 1% and 10% of the products. Above all, we can confirm that our LR reputation system has some degree of similarity on 1% and 10% top ranked products, but this degree declined afterwards. The stability of the results over the three datasets confirm that our LR model significantly produces different results and better accuracy as confirmed by MAE.

Table 6. Ranking of models based on MAE over three datasets

| Rank | 100K | 1M | 10M |
|---|---|---|---|
| 1 | RT | RT | RT |
| 2 | LR | LR | LR |
| 3 | KNN | KNN | KNN |
| 4 | SVR | SVR | SVR |
| 5 | Median | Median | Median |
| 6 | BetaDR | BetaDR | BetaDR |
| 7 | Dirichlet | Dirichlet | Average |
| 8 | Bayesian | Bayesian | Dirichlet |
| 9 | Average | Average | Bayesian |
| 10 | IMDb | Fuzzy | Fuzzy |
| 11 | Fuzzy | IMDb | IMDb |
| 12 | LQ | LQ | LQ |

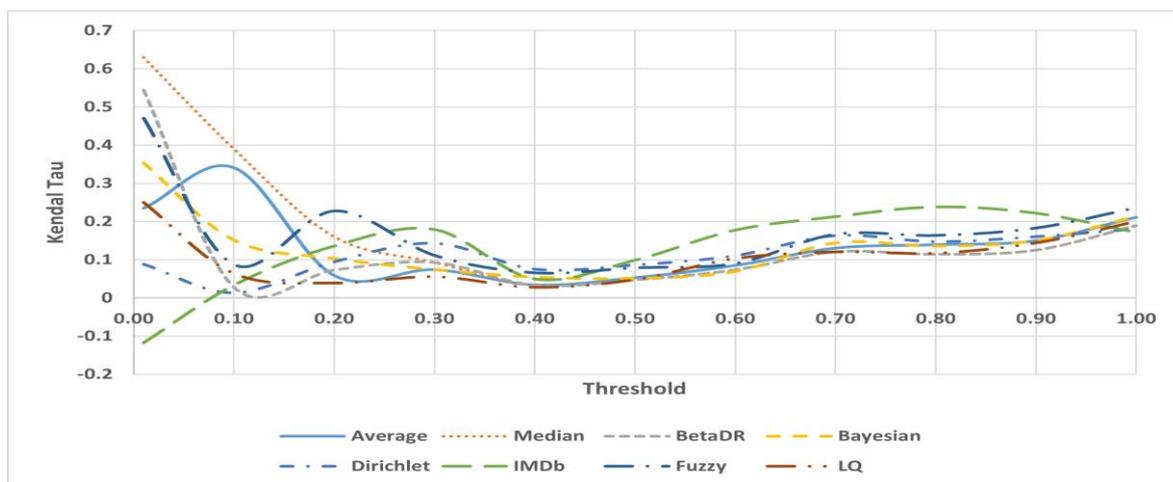

(a) 100K dataset.

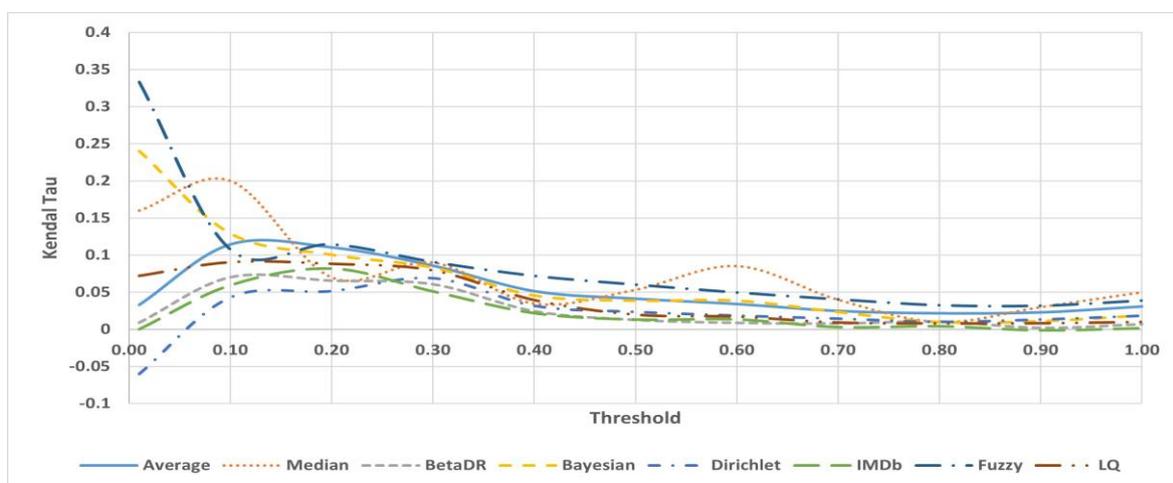

(b) 1M dataset.

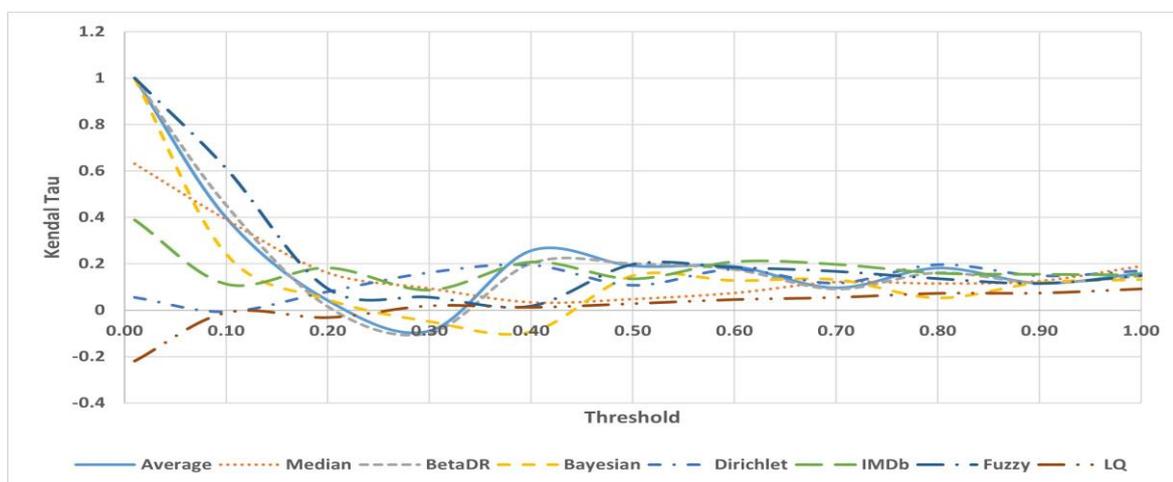

(c) 10M dataset.

Fig. 2. Kendall Tau Coefficient comparison of LR and previous reputation systems over employed datasets

Regarding RT model, we can notice that our model and the three models (Fuzzy, Bayesian and BetaDR) rank only top 1% and 10% products similarly on 100K dataset as shown in Fig. 3(a), but they decline after using 10%, which confirms that ranking lists are independent from each other. The good point here is that all similarity lines decline to reach near to zero after 20% which tell us that the RT model produces different reputation scores than other models. For other comparisons over 1M and 10M datasets we observe relatively the same trend that our model ranks top products differently from other reputation systems as shown in Figures 3(b) and 3(c). In summary, we can figure out that the ranking order of the top 10% of product list generated by our model is relatively different from other reputation systems, over three datasets.

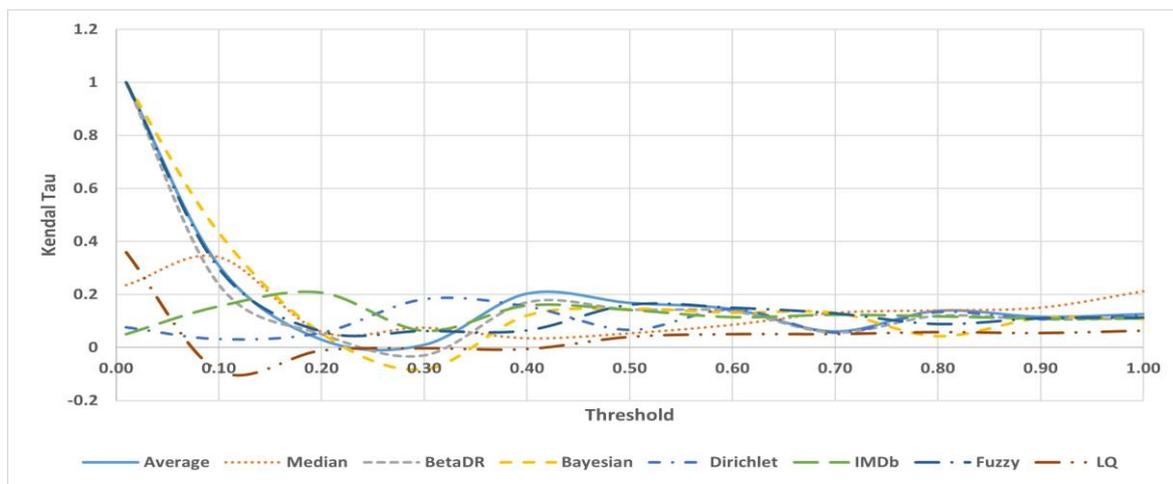

(a) 100K dataset

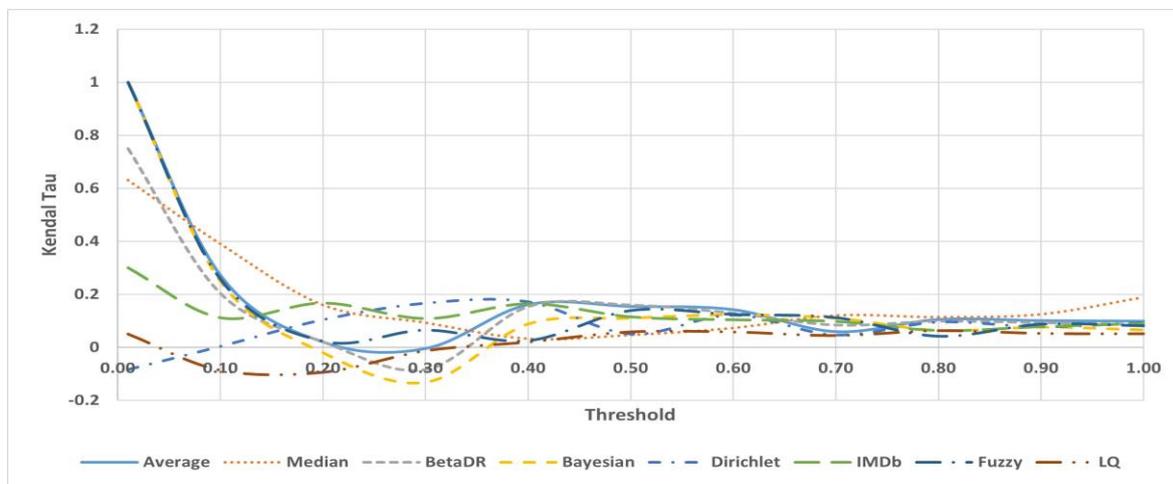

(b) 1M dataset.

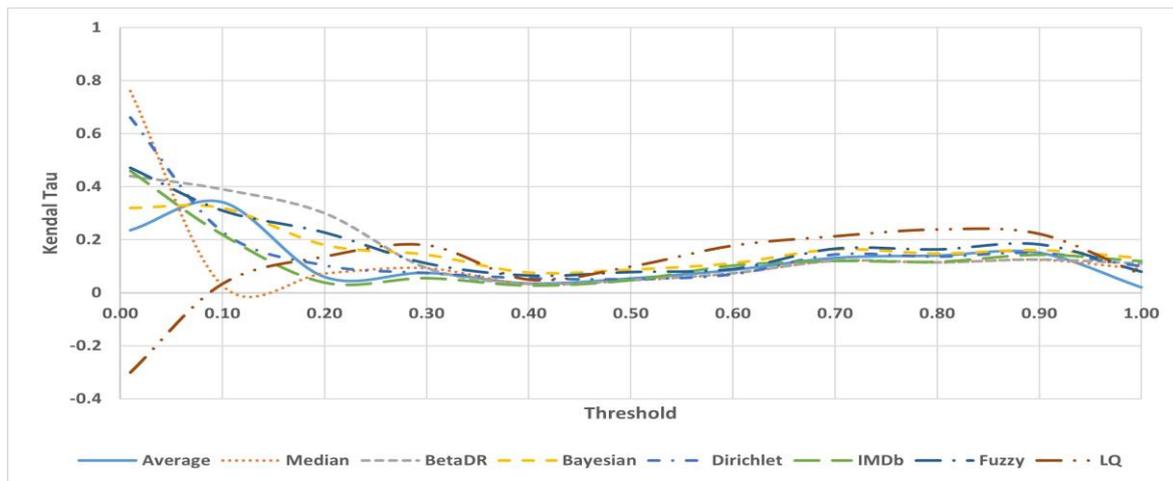

(c) 10M dataset.

Fig. 3. Kendall Tau Coefficient comparison of RT and previous reputation systems over employed datasets

Fig. 4 shows comparison between KNN reputation system and other models over three datasets. For 100K dataset as shown in Fig. 4(a), it is noticed that our model ranks 1% and 10% of top product relatively similarly to Median and Bayesian models. However, the correlation degree began to decline after using top 20%. The main observation here is that there is no stable relation with Fuzzy model. The trend is slightly different over 1M as shown in Fig. 4(b) where our model shows relatively small similarity degree with other models at 1% and 10% which ranks top products quite similarly to BetaDR, Average and Median. Notably, our model and LQ, and IMDb models rank top products differently, which indicates that our model is more accurate as confirmed by MAE. For Large dataset 10M we can notice that our model produces quite similar top product list to Fuzzy, LQ and BetaDR when we look at top 1% and 10% of the products. Finally, we can confirm that our RT reputation system has some degree of similarity on 1% and 10% top ranked products, but this degree declined afterwards. The stability of the results over the three datasets confirm that our KNN model significantly produces different results and better accuracy as confirmed by MAE.

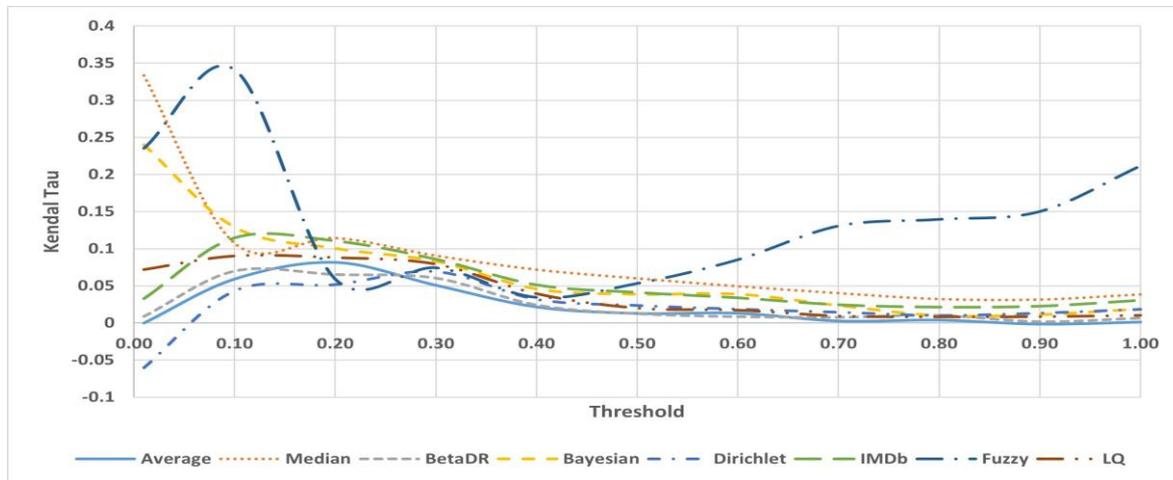

(a) 100K dataset.

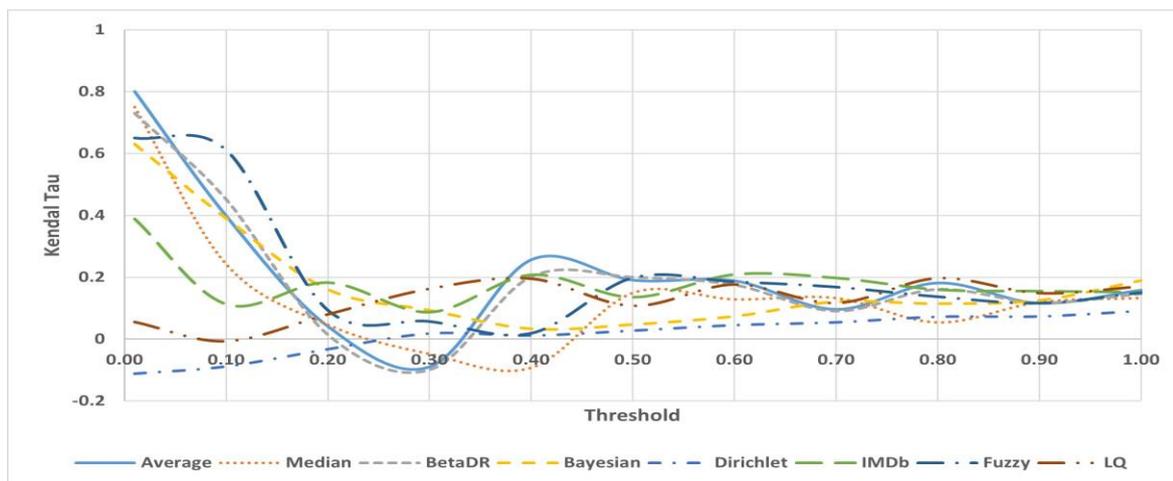

(b) 1M dataset.

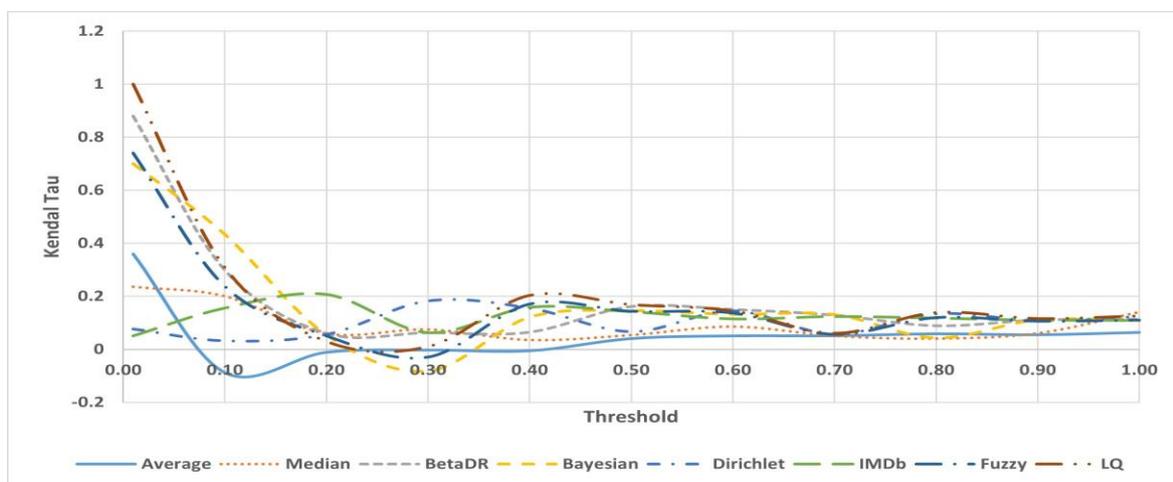

(c) 10M dataset.

Fig. 4. Kendall Tau Coefficient comparison of KNN and previous reputation systems over employed datasets

*International Journal of Automation and Computing*

Fig. 5 shows comparison between SVR reputation system and other models over three datasets. For 100K dataset (Fig. 5(a)), we can notice that our model ranks 1% and 10% of top product similar to Fuzzy model and quite similar to Median and BetaDR. However, the correlation degree began to decline after using top 30%. For 1M (Fig. 5(b)), the trend is similar where our model shows relatively similarity degree with Average and Bayesian at 1% and 10% which ranks top products quite similarly. Regarding 10M dataset (Fig. 5(c)) we can notice that our model produces quite similar top product list to Fuzzy when we look at top 1% and 10% of the products. Finally, we can confirm that our SVR reputation system has some degree of similarity on 1% and 10% top ranked products, but this degree declined afterwards.

Finally, we revise the proposed research questions:

**RQ1**: Does the extract variables have great effect on computing consumer trust?
Ans. Yes, according to MAE and Kendall Tau results, the extracted variables have capability to help in predicting consumer reliability based on the employed machine learning methods. Our models with four factors give high accuracy in comparison with other reputation systems that depend on one or two factors.

**RQ2**: Does using machine learning enables us to compute consumer trust efficiently and thus enhance accuracy of rating aggregation?
Ans. According to MAE validation method we noticed that all results, over all data sets, are quite small which means that our reputation systems have capability to predict the correct weight for each consumer based on its provided ratings.

**RQ3**: Which machine learning method can produce better performance?
Ans. According to MAE validation results we notice that RT Machine learning model gives higher accuracy over the three employed datasets.

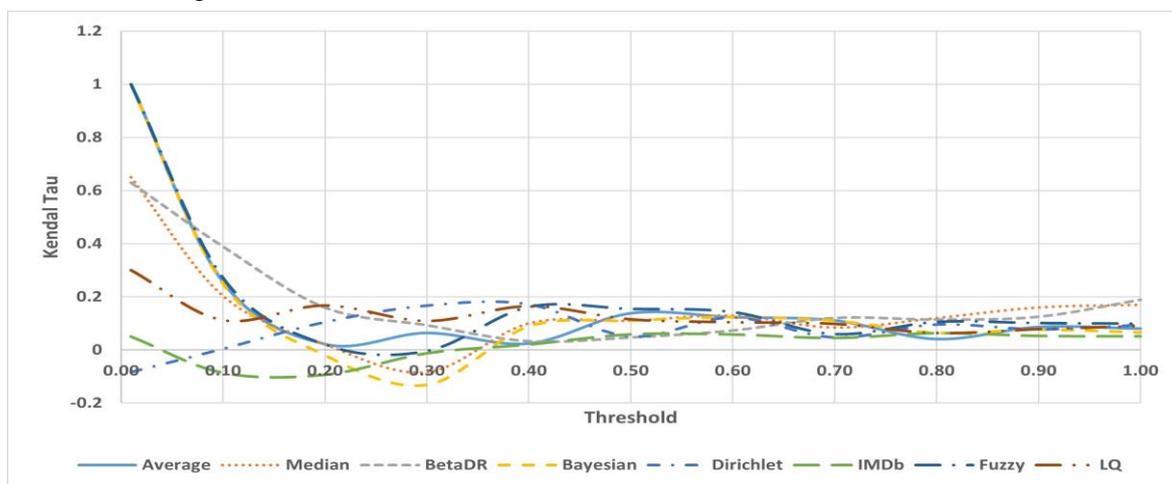

(a)   100K dataset

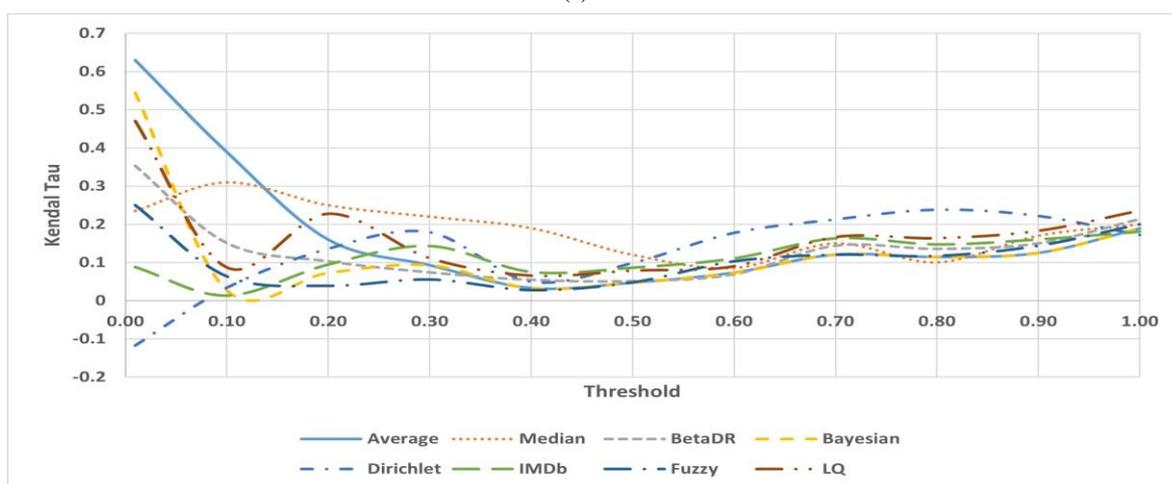

(b)   1M dataset.

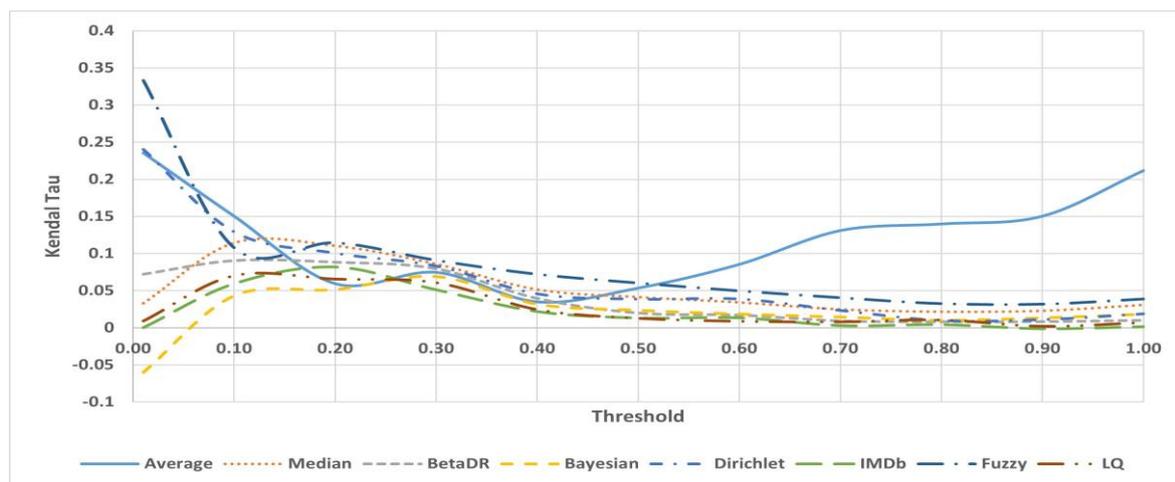

(c) 10M dataset.

Fig. 5. Kendall Tau Coefficient comparison of SVR and previous reputation systems over employed datasets

## 5 Conclusions

With increasing popularity of online shopping markets, reputation systems emerged as a solution to facilitate their choices decision. This paper proposed a new reputation system based on weighted average approach. The weight is predicted from consumer profile data using machine learning algorithms. In fact, four machine learning algorithms were used to predict consumer weights. These weights are used within weighted average model to compute product reputation score. To predict weights, we constructed a new consumer profile matrix that consists of six variables: number of positive ratings, number of neutral ratings, number of negative ratings, fluctuation, experience and reliability. In this approach we focused on giving higher weights for high trusted consumers. We believe that rating weights should relate to the reliability of rating given by a consumer, as the this reflects how end user view an item.

The constructed reputation models have been evaluated against various reputation models from literature. The results showed that the proposed approach surpasses all previous models over MovieLens datasets using MAE evaluation measure. According to the MAE validation method, we concluded that all results, over all data sets, are quite small. In more detail, the proposed approach performs significantly better than all other models by reducing the error generated in rating predictions. Also, we noticed that the proposed approach produces a relatively different ranking for items based on the reputation scores compared with the naive and baseline methods. Besides, it provides a different ranking compared with the other sophisticated models such as LQ and Dirichlet. This indicates the significance of proposing the new reputation model based on machine learning and addresses the need to evaluate reputation models with regard to the accuracy of the ranked items list, which was performed in the second part of the experiment. According to Kendall tau coefficient validation method, the overall results show the same trends in all figures. These results demonstrate that our proposed approach produces relatively different ranked product lists than previous models, which necessarily confirm that our models are more accurate based on MAE. These encouraging results have subsequent implications on recommender systems when they are integrated with our proposed approach. This kind of integration is supposed to provide better purchasing recommendations and facilitate user experience on online shopping markets. However, there is still a need to investigate this issue with state of art recommendation systems.

## Acknowledgements

The authors are grateful to the Applied Science Private University, Amman, Jordan, for the financial support granted to cover the publication fee of this research article.

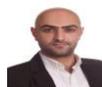

**Ahmad Alqawadri** is Master student pursuing MSc in computer Science at Applied Science Private University. His research interests include Machine learning and data mining.

E-mail: ahmad.qawadri@asu.edu.jo

ORCID iD: 0000-0000-0000-0000

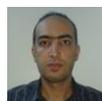

**Mohammad Azzeh** is a full professor at the Department of Computer Science in the Faculty of Information Technology at Applied Science University. He earned his PhD in Computing from University of Bradford in 2010, Bradford, UK. M.S.C in Software Engineering from University of the West of England, Bristol, UK. Dr. Mohammad has published over 50 research articles in reputable journals and conferences such as IET Software, Software: Evolution & Process, Empirical Software Engineering and Systems & Software. His research interests focus on Software Cost Estimation, Empirical Software Engineering, Data Science, Mining Software Repositories, Machine Learning for Software Engineering Problems. Dr. Mohammad was Conference chair of CSIT2016 and CSIT2018, and he is co-chair of many IT-related workshops.

E-mail: m.y.azzeh@asu.edu.jo (Corresponding author)

ORCID iD: 0000-0002-0323-6452

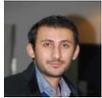
**Fadi Almasalha** is an Associate Professor at the Faculty of Information Technology at Applied Science Private University in Amman, Jordan; received his M.Sc. in computer science from New York Institute of Technology, in 2005 and Ph.D.in Computer Science from University of Illinois at Chicago, in 2011. In fall of 2011, he joined the Department of Computer Science at the Applied Science University. Dr. Fadi Almasalha received his Associate rank in 2016, during his appointment as the head of computer science department. Dr. Fadi has published more than 10 technical papers, journals and book chapters in refereed conferences and journals in the areas of multimedia systems, data mining, and cryptography.

E-mail: f_masalha@asu.edu.jo

ORCID iD: 0000-0002-7900-2409